\documentclass[12pt]{article}
\usepackage[utf8]{inputenc}
\usepackage{times}
\usepackage{fullpage}
\usepackage{eufrak}
\usepackage{amssymb}
\usepackage{amsmath}
\usepackage{amsthm}
\usepackage{tikz}
\usepackage{graphicx}
\usepackage{booktabs} 
\usepackage{float}
\usepackage{pgfplots}
\usetikzlibrary{arrows.meta}
\usepackage{xcolor}

\newcommand{\indep}{\raisebox{0.05em}{\rotatebox[origin=c]{90}{$\models$}}}

\pgfplotsset{compat=1.14}

\makeatletter
\newcommand{\printfnsymbol}[1]{%
  \textsuperscript{\@fnsymbol{#1}}%
}
\makeatother
\begin{document}
\title{
{Unifying Causal Models with Trek Rules}}

\author{Shuyan Wang \\
  Department of Philosophy \\
  Carnegie Mellon University\\
  \texttt{shuyanw@andrew.cmu.edu}}

\maketitle

\begin{abstract}
In many scientific contexts, different investigators experiment with or observe different variables with data from a domain in which the distinct variable sets might well be related. This sort of fragmentation sometimes occurs in molecular biology, whether in studies of RNA expression or studies of protein interaction, and it is common in the social sciences. Models are built on the diverse data sets, but combining them can provide a more unified account of the causal processes in the domain. On the other hand, this problem is made challenging by the fact that a variable in one data set may influence variables in another although neither data set contains all of the variables involved.  

Several authors have proposed using conditional independence properties of fragmentary (marginal) data collections to form unified causal explanations when it is assumed that the data have a common causal explanation but cannot be merged to form a unified dataset.  These methods typically return a large number of alternative causal models.  The first part of the thesis shows that marginal datasets contain extra information that can be used to reduce the number of possible models, in some cases yielding a unique model.

\end{abstract}

\section{Introduction}
Methods for unifying theories are typically particular to the theories, and depend on some deep insight into a shared fundamental structure. In contrast, for simple causal models, which abound in the biomedical and social sciences, general procedures for unification have been proposed. Causal relations between variables can be discovered by randomized experiments or by other interventions and also by analyzing non-experimental data.  Many algorithms have been designed to find causal relations between variables from datasets. Most of these algorithms search for causal relations for variables measured in one dataset and output a directed graph in which variables directly connected by an edge are hypothesized to have relatively direct causal relations.  However, due to restrictions such as time, location or privacy, all of the variables participating in a causal mechanism may not be measured jointly, in which case researchers may have several datasets sharing some but not all variables -- overlapping variable sets. 

Such marginal datasets impose restrictions on identifying causal relations, since interactions between some variables are not observed.  Using only marginal datasets, even if researchers know that variables from all these datasets are from a shared causal system, there may be too many possibilities for an informative estimation of causal relations\cite{NIPS2008_3604}. Concatenating datasets, then running algorithms which can work with missing variables could handle this problem, but this method requires strong assumptions on how and why the values are missing and is not feasible for confidential data that cannot or will not be shared. 

Besides concatenating datasets, several other responses have been made to this problem.  The ION algorithm \cite{10.1093/bjps/axi146}\cite{NIPS2008_3604} takes as input a package of partial causal graphs, which are generated by running algorithms allowing for “latent variables” (e.g., the Fast Causal Inference (FCI) algorithm \cite{book}) on each of the marginal datasets, and returns a package of unified graphs, all of which are acyclic, contain all variables, and are consistent with the conditional independence and dependence information estimated from the input data \cite{NIPS2008_3604}.  ION gives a set of possible causal mechanisms between all variables measured in any of the datasets.  Integrative Causal Analysis (INCA) has also used conditional independence relations in analyzing data over different variable sets to generate causal models that are consistent with all marginal datasets\cite{Tsamardinos2012TowardsIC}. 

The methods mentioned above work by finding unified causal graphs that include variables in each marginal dataset while preserving all the marginal conditional (in)dependence relations.  The basic idea is to find those unified models that can account for all of the conditional independence and dependence relations found in the marginal data sets. Assuming the well-known Causal Markov Condition and Faithfulness assumption \cite{book}, this procedure can be revealing.  For example, suppose the true causal relations are given by the graph in Figure \ref{fig:figure1}:

\begin{figure}[H]
    \centering
    \includegraphics[width=100mm]{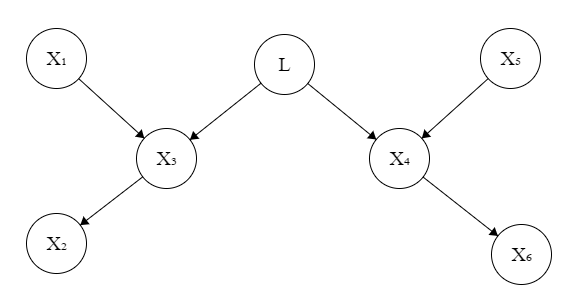}
    \caption{}
    \label{fig:figure1}
\end{figure}

Suppose the observed data sets are for $\{X_1, X_3, X_4\}$ and $\{X_4, X_5, X_6\}$. L is not observed in any data set.  From sufficiently large samples, conditional independence methods can recover Figure 1 uniquely. In other cases, however, even apparently simple cases, the methods returns a plethora of alternative causal structures. For example, for the structure

\begin{figure}[H]
    \centering
    \includegraphics[width=\linewidth]{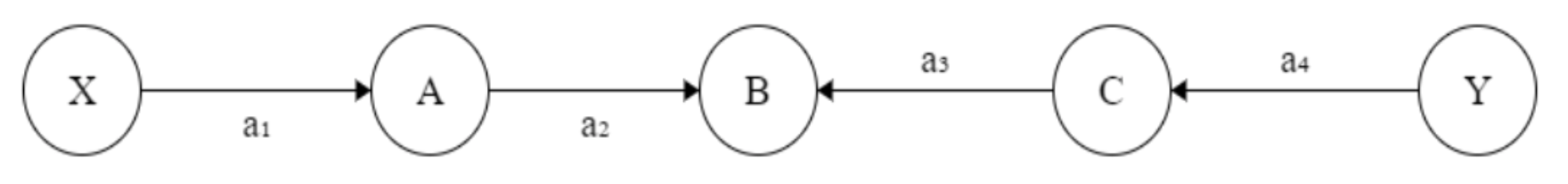}
    \caption{}
    \label{fig:figure2}
\end{figure}
with marginal datasets:
\begin{itemize}
    \item $\{X, Y, A\}$
    \item $\{X, Y, B\}$
    \item $\{X, Y, C\}$
\end{itemize}
five distinct structures can account for the marginal conditional independence and dependence relations. In many cases the number of alternative unifying models is very large.

However, marginal dependence or conditional independence relations are not the only information that can be used from marginal datasets.  We will show that if the relationships are linear, TREK rules can aid in estimating the causal connections between two variables that only appear in separate datasets.  Here, we explore the use of marginal correlations with the TREK rules to estimate a unified model, with results that can be more informative than those obtained solely from marginal conditional dependence and independence relations in multiple datasets.

\section{The TREK Rule}

We use directed graphs to represent causal relations between variables; each node represents one variable, each edge represents one (relatively) direct causal relation, with the direction from the cause (parent) to the effect (child).  A node $Y$ is called a descendant of another $X$ if $Y$ can be reached by following a directed path starting from $X$, which is called an ancestor of $Y$.  We assume the joint probability distribution on the variables respects the Markov condition, i.e., all variables conditioned on their parents are independent from the set of all of their non-descendants.  Two acyclic directed graphs (DAG) are called Markov equivalent if they entail the same conditional independence relation based on the Markov condition. The Faithfulness assumption, that all conditional independence relations are consequences of the Markov condition is made but in some cases is not necessary.

We  assume that all causal relations are linear: as an effect, every variable is a linear combination of influences from its direct causes, included unmeasured “disturbances” of each variable that are independent of its measured causes. We also assume that no measured variable is a deterministic function of any set of other measured variables.  Formally, if we use linear coefficient a to show the strength of the influence from one direct cause, $C$, to an effect $E$, and assume that $E$ is also caused by some other unobserved noises $e$ independent from the direct cause $C$, the relation between $C$ and $E$ is:

\begin{center}
    $E =  aC+e$  
\end{center}  
If $E$ has more than one direct cause, the relation between them is;

\begin{center}
    $E =\Sigma_{i=1}  a_iC_i+e$  
\end{center}  
The TREK rule\cite{book} can be derived from these assumptions.  A trek between two variables, $X$ and $Y$, is defined as either a directed path starting from one variable that ends at another, or two directed paths starting from a common third variable  $Z$, the two paths intersecting only at $Z$, with one path ending at $X$ and another at $ Y$.  The following figure shows these two types of treks.  In the top trek, $X_0$ is a remote cause of $X_n$; in the bottom trek, $X_0$ and $X_n$ are indirect effects, i.e. descendants, of a common cause $X_{23}$.  

\begin{figure}[H]
    \centering
    \includegraphics[width=100mm]{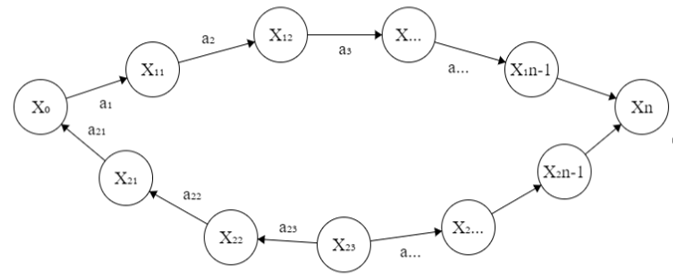}
    \caption{}
    \label{fig:figure3}
\end{figure}
 
The TREK rule says:  the correlation between any two standardized variables is the sum of products of linear coefficient on each trek between them.  “Standardized” means that all variables are rescaled to have a mean $(\mu)$ of 0 and variance 1.  For example, in the figure \ref{fig:figure3}, the correlation ($\rho$) between (standardized) $X_0$ and $X_n$ is:
\begin{center}
    $\rho_{X_0,X_n}=a_{1}a_{2}a_{3}...a_{n}+a_{21}a_{22}a_{23}...a_{2n}$
\end{center}

Now we are going to prove the TREK rule for linear, acyclic systems of standardized variables with independent disturbance terms.\color{white}
\begin{proof}
first line.

   \color{black}
   \textit{Proof sketch.}
   $X$ is the standardization of $X'$ iff $X = \dfrac{(X' - \mu _X' )}{ \sqrt{VAR_{X'}}}$. The mean of any standardized variable, $X$, is 1 and the variance, $VAR(X)$, is 1. The correlations of two standardized variables, $X$, $Y$, is the expectation of their product, $E(XY)$.

A trek is a pair of directed paths terminating in two distinct variables, $X$, $Y $and intersecting at a single variable, $St$, the source of trek $ t$.  Or, a single directed path from $X$ into $ Y$, in which case $St = X$.  

\textbf{Notation:} $a_{it}$,  $b_{it}$, $c_{it}$, etc. denote the coefficient for the ith edge in trek $t $ starting from the source.

\textbf{Remark 1:} $X \rightarrow Y$ denotes $Y = aX +e_Y$. If $X \rightarrow Y$ is the graph, the correlation of $X$, $Y$ is $E(XY) = E(X (aX + e_Y)) = E(aX^2 + aXe_Y) = a + 0 = a$, because  $X$ and $e_Y$ are uncorrelated. Suppose for every causal graph of length $m$, the correlation of the terminal variables, $X$, $Y$ is $ a_1...a_m$.  Let $G$ be a chain graph of length $m+1$ by adding one edge $Y \rightarrow Y_{m+1}$. Then $E(XY_{m+1}) = E(X (a_{m+1}Y + e_{Y_{m+1}})) =a_{m+1}E(XY) + E(Xe_{Y_{m+1}})= a_1 . .. a_ma_{m+1} = \rho_{XY_{m+1}}$. Using an induction argument, we conclude that the correlation of a causal chain of any length is given by the product of the edge coefficients. 

\textbf{Remark 2:} $ X \leftarrow Z \rightarrow Y$ is the graph of the linear system $X = aZ + e_X; Y = bZ + e_Y$.  $E(XY) = E[(aZ + e_X)(bZ + e_Y)] = E(abZ^2) = ab E(Z^2)  = ab = \rho_{XY}$ since $E(Z^2) = Var(Z) =1$. Applying Remark 1 to each side of $ X \leftarrow Z \rightarrow Y$, for any pair of directed paths$ Z \rightarrow X_1 \rightarrow\dots  \rightarrow X_n$ and$ Z \rightarrow Y_1\rightarrow\dots\rightarrow Y_m$, with respective edge coefficients $a_1,\dots ,a_n$ and $b_1,\dots b_m$, $E(X_nY_m) = a_1\dots a_n b_1\dots b_m Var(Z) =  a_1\dots a_n b_1\dots b_m  = \rho_{X_nY_m}$.

Consider the graph $ X \leftarrow Z \rightarrow Y$ again, if we add an additional trek between $X$ and $Y$, such that the length of the path between the source $St$ and either $X$ or $Y$ does not exceed 1, it is easy to see that $E(XY) = ab+c_Xc_Y = \rho_{XY}$ where $c_N$ is the coefficient of the edge connecting $St$ and $N$ or the correlation between $St$ and $N$. By an induction, for $X$ and $Y$ connected by any two treks $t_1$ and $t_2$  with coefficients $a_i$ and $b_i$, $E(XY) = \Pi_i a_i VAR(St_1) + \Pi_i b_i VAR(St_2) = \Pi a_i + \Pi b_i  = \rho_{XY}.$ By an induction, always, $\rho_{XY} = \Sigma_t \Pi_i c_{it} $, which is the TREK rule.
\end{proof}
\color{black}

\section{Estimating Causal Connections Using the TREK Rule: Examples}
In this section we give three examples to show how the TREK rule can be used either to estimate unified causal graphs formed by variables measured in marginal datasets or to reveal information that is not explicit when analyzing causal connections based only on conditional independence.  Each example starts with a true causal graph and marginal datasets.  Assuming faithfulness, linear relations and Gaussian distributions, we examine what dependence and independence relation can be obtained from these datasets.  Based on the obtained (in)dependence and correlations measured from these datasets, the TREK rule helps to estimate causal connections and narrow down the range of possible unified causal graphs.
\subsection{Case One\cite{Danks2017AmalgamatingEO}}

The true causal graph is:

\begin{figure}[H]
    \centering
    \includegraphics[width=100mm]{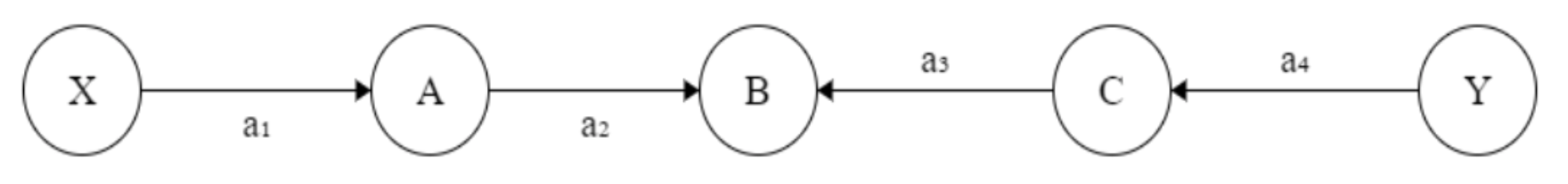}
    \caption{True Graph in Case One}
    \label{fig:figure1n}
\end{figure}

with marginal datasets:
\begin{itemize}
    \item $\{X, Y, A\}$
    \item $\{X, Y, B\}$
    \item $\{X, Y, C\}$
\end{itemize}
From the true causal graph we know that the independence relations we get from the three datasets above are:
\begin{itemize}
    \item $X\indep Y$
    \item $A\indep Y$
    \item $X\indep C$
\end{itemize}
Based on faithfulness we can tell that $B$ is a collider with $X$ and $Y$ on each side, because $X $ and $Y$ become dependent conditioning on $B$.  We also know there is no trek connecting $A$ and $Y$ because they are marginally independent.  Similarly, there is no trek connecting $X$ and $C$.  All the graphs below agree with these dependence and independence relations:
\begin{figure}[H]
    \centering
    \includegraphics[width=100mm]{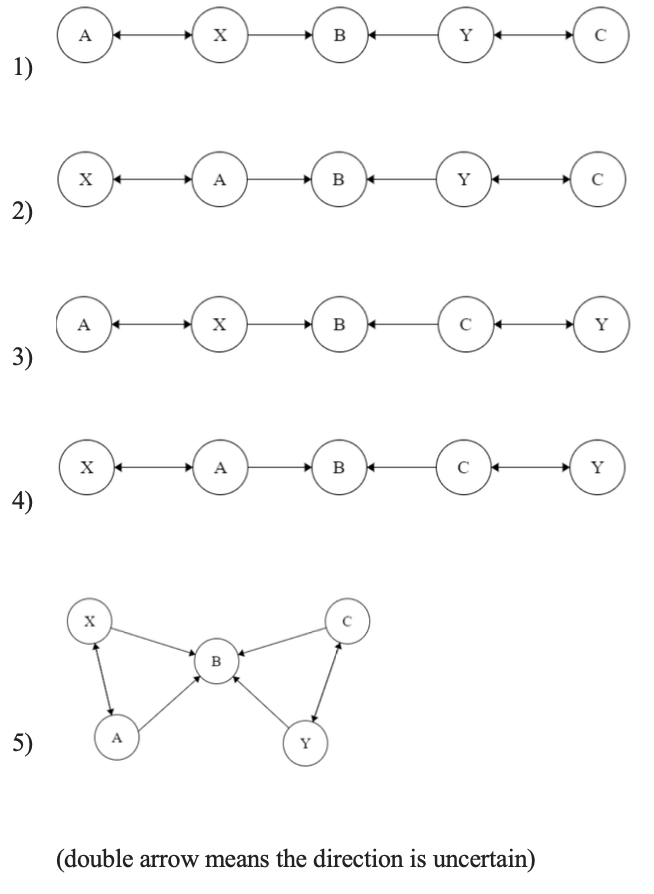}
    \caption{Possible Graphs}
    \label{fig:figure4}
\end{figure}

To rule out some of these candidates, we have to use more than independence and conditional independence information.  One choice is correlation.  The non-zero correlations we know are $\rho_{XA}$, $\rho_{XB}$, $\rho_{YC}$ and $\rho_{YB}$.  By comparing some of these correlations, we can rule out any causal graph such that if this graph were true, the TREK rule would be violated.  For instance, if the true graph is 2) or 4), by TREK rule:
\begin{center}
    $\rho_{XB}=\rho_{XA}\rho_{AB}$
\end{center}
Since $|\rho_{AB}|\leq1$, we have $|\rho_{XB}|\leq|\rho_{XA}|$.

That is to say, if $|\rho_{XA}|$ is smaller than $|\rho_{XB}|$, the true graph cannot be 2) or 4).  Similarly, comparing the absolute value between $\rho_{YC}$ and $\rho_{YB}$ may rule out 3) and 4).  The effect of applying TREK rules in this way is summarized in the table below (``X” means ``the condition in the row rules out the model in the column”):
\begin{figure}[H]
    \centering
    \includegraphics[width=\linewidth]{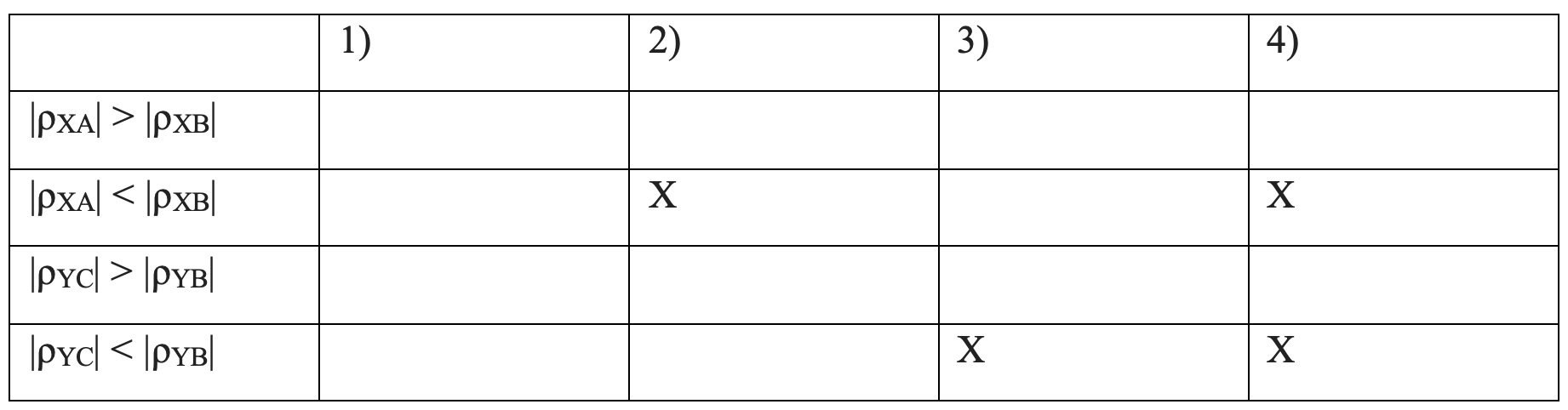}
    \caption{}
    \label{fig:figure5}
\end{figure}
If we can measure either $\{A, B\}$ or $\{C, B\}$, we can compare $\rho_{BA}$ and $\rho_{XB}$ or $\rho_{BC}$ and $\rho_{YB}$ and rule out more cases.  For instance, if we know that $|\rho_{BA}|$ is greater than $|\rho_{XB}|$, we can rule out graph 1) and 2). 
\subsection{Case Two}
\begin{figure}[H]
    \centering
    \includegraphics[width=100mm]{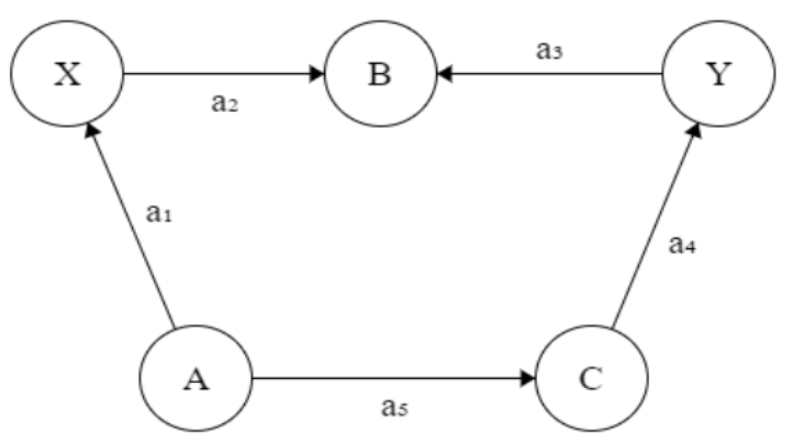}
    \caption{True Graph for Case Two}
    \label{fig:figure6}
\end{figure}
The measured datasets are:
\begin{itemize}
    \item $\{X, Y, A\}$
    \item $\{X, Y, B\}$
    \item $\{X, Y, C\}$
\end{itemize}
The independence relations we can get from those datasets are:
\begin{itemize}
    \item $X\indep Y|A$
    \item $X\indep Y|C$
\end{itemize}

From these conditional independences, we know that every trek connecting $X$ and $Y$ contains $A$ and $C$.  We can also determine the relative position of $A$ and $C$ in the trek:  the correlation with $X$ and the variable closer to $X$ has a larger absolute value.  Similar to Case 1, the TREK rule yields:  
\begin{center}
    $|\rho_{XA}|= |\rho_{AC}||\rho_{XC}|$ if $C$ is between $X$ and $A$
\end{center}
or
\begin{center}
    $|\rho_{XC}|= |\rho_{AC}||\rho_{XA}|$ if $A$ is between $X$ and $C$
\end{center}

Since the absolute value of correlation is between 0 and 1, comparing absolute values of correlations can reveal the causal connection between these three variables.
Since $X$ and $Y$ are not independent conditioning on $B$, we know that unlike $A$ and $C$, $B$ is not in every trek connecting $X$ and $Y$.  Therefore, $B$ maybe a collider.  Note that from the dataset $\{X, Y, B\}$, we should find that the marginal correlation between any pair of variables is different from the conditional correlation (i.e., $\rho_{XY|B} \neq \rho_{XY}$, $\rho_{XB|Y} \neq \rho_{XB}$, $\rho_{YB|X} \neq \rho_{YB}$).  This means that each variable in this dataset is either in a trek connecting the other two variables, or a collider or descendant of a collider in path connecting the two variables.  If $B$ is a collider, this could only happen when neither $A$ nor $C$ is in the $X-B-Y$ path.  Therefore, since $B$ is a collider, we end up with figure \ref{fig:onePossibleGraph}:
\begin{figure}[H]
    \centering
    \includegraphics[width=100mm]{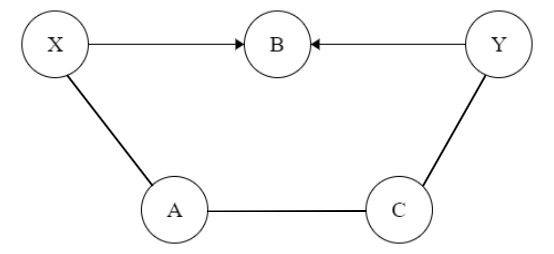}
    \caption{Possible Graph for Case Two}
    \label{fig:onePossibleGraph}
\end{figure}
In this case, we can recover the full causal graph except for the directions of the edges between $X$, $A$, $C$ and $Y$ Comparing this case with case 1, we see that they have the same marginal datasets and the only difference between them is that in case 2 $A$ and $C$ are connected.  This extra edge reduces the number of candidates for the true causal graph (up to Markov indistinguishability) from five to one.  That is because the direct connection between $A$ and $C$ enables $X$ and $Y$ to be connected by a trek, the longest trek in the true graph.  From the marginal datasets we see that every variable in this trek is measured together with the endpoints ($X$ and $Y$) of this trek, which enables us to determine the exact structure of this trek.
If $B$ is not a collider, $B$ is in at least one trek connecting $X$ and $Y$. As stated above, $A$ and $C$ should also be in the trek that contains $B$.  Since the causal graph is assumed to be a DAG, for the set $\{X, Y, B\}$ either $X$ or $Y$ has to be a descendent of collider in a path connecting the other two variables.  The only two situations compatible with the ``$\rho_{XY|B} \neq \rho_{XY}$, $\rho_{XB|Y} \neq \rho_{XB}$, $\rho_{YB|X} \neq \rho_{YB}$” information are figure \ref{fig:figure8} (i) and (ii) \footnote{If $B$ directly connects to $A$ and $C$, either $\rho_{XB|Y}\neq\rho_{XB}$ or $\rho_{YB|X} \neq \rho_{YB}$ is violated}.  

\begin{figure}[H]
    \centering
    \includegraphics[width=\linewidth]{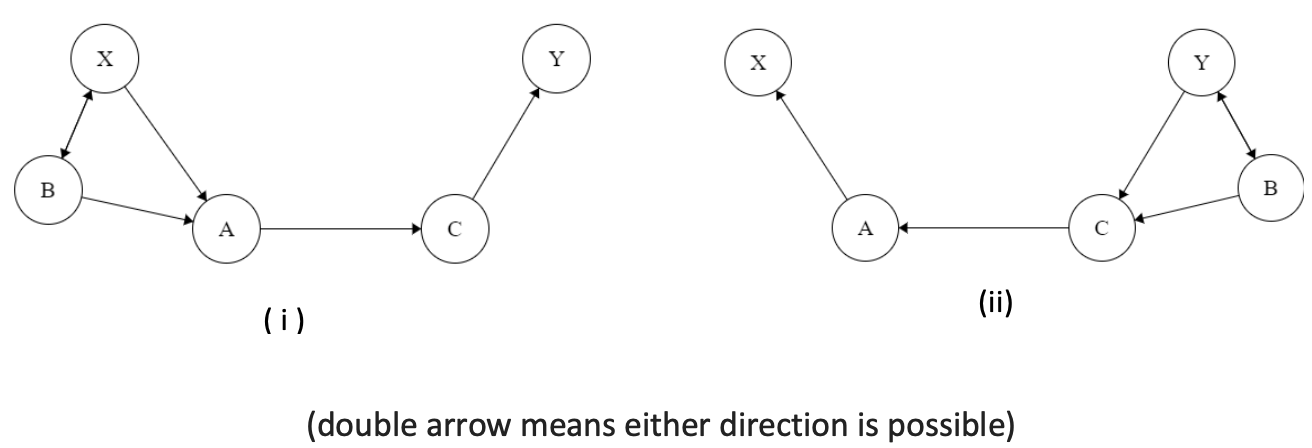}
    \caption{Possible Graph for Case Two}
    \label{fig:figure8}
\end{figure}
Therefore, in this case, we can narrow down the possible true graph into three situations (figure \ref{fig:onePossibleGraph} and \ref{fig:figure8}) using the inequality between different correlations entailed by the TREK rule.

\subsubsection{Case Three}
This case shows that the TREK rule can be used to check for the existence of latent variable or directed edge.  Only one graph is used here for illustration, but this method is at least theoretically available to graphs with this 4(5)-member subgraph.  Suppose it is known that there is no direct connection between $X_2$, $X_3$ and the true, unknown model is as shown in figure \ref{fig:Case3T}: \begin{figure}[H]
    \centering
    \includegraphics[width=100mm]{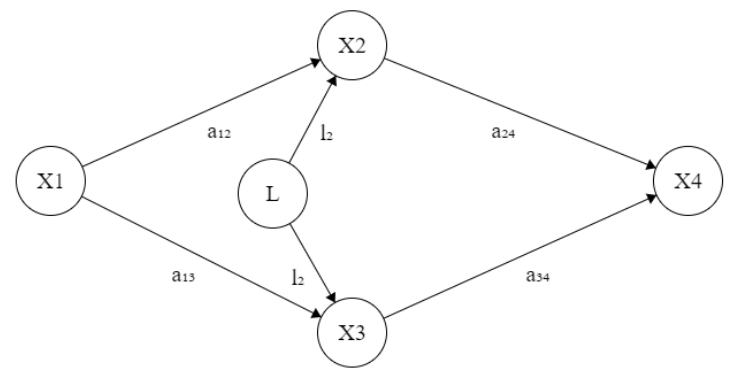}
    \caption{True Graph for Case Three}
    \label{fig:Case3T}
\end{figure}
In the figure \ref{fig:Case3T}, $L$ denotes a latent variable.  If all four observed variables, $X_1$ to $X_4$, can be measured together, whether $L$ exists or not will result in different conditional independence relations, in which case the method introduced here is redundant.  If, however, only some of those variables can be measured, for instance {X1, X2, X4} and {X1, X3, X4}, then whether L exists is not obvious anymore.  The method introduced here can be used to check the existence of L in some of the possible graphs.
If the latent variable does not exist, the true graph could be:
\begin{figure}[H]
    \centering
    \includegraphics[width=100mm]{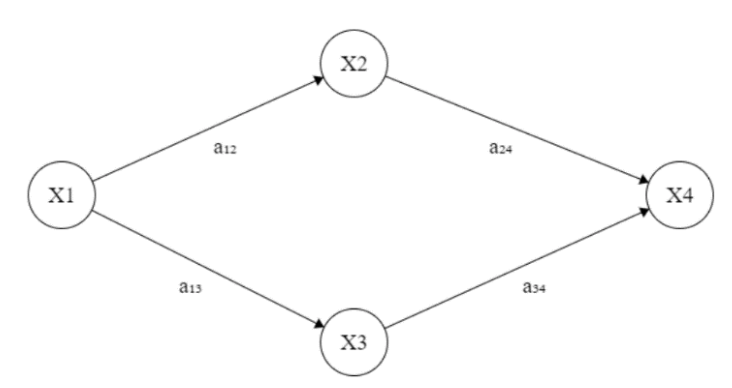}
    \caption{Possible Graph for Case Three}
    \label{fig:Case3P}
\end{figure}
By TREK rule, we have:
\begin{enumerate}
    \item $\rho_{12}=a_{12}$,$\rho_{13}=a_{13}$
    \item $\rho_{24}=a_{12}a_{13}a_{34}+a_{24}$
    \item $\rho_{34}=a_{12}a_{13}a_{24}+a_{34}$
    \item $\rho_{14}=a_{12}a_{24}+a_{13}a_{34}$
\end{enumerate}
Since all the correlations needed are contained in the two marginal datasets, we can get $a_{24}$ and $a_{34}$ by solving 1), 2) and 3); then by 1), we can check whether the equation 4) holds or not.  If the equation holds, the latent variable does not exist. If the equation does not hold, then there should be extra connection between $X_2$ and $X_3$, which could be a latent common cause of $X_2$ and $X_3$ or a direct connection between them.

\section{Case with Non-Gaussianity}
If Gaussian distributions are assumed, causal relations can only be estimated up to the Markov equivalence class and we may not know the direction of many edges in a causal graph.  However, if we assume non-Gaussian distributions, we can apply algorithms, such as LiNGAM, to each marginal dataset, which return partial graphs where each edge has a direction unless a latent common cause exists \cite{Shimizu:2006:LNA:1248547.1248619}.  In the non-Gaussian case, we can determine whether a path between two variables is a trek.  If directions of edges tell us that a path is a trek, we can apply the TREK rule directly to estimate the connection between the terminal variables of the trek. Consider a case where the causal graph is figure \ref{fig:Case3P2}: 
\begin{figure}[H]
    \centering
    \includegraphics[width=100mm]{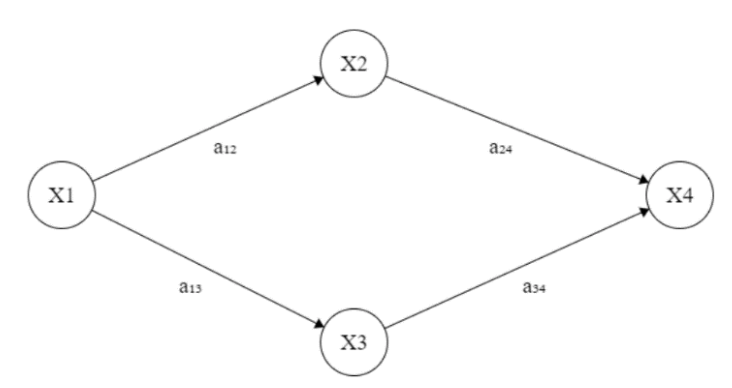}
    \caption{True Graph with non-Gaussianity}
    \label{fig:Case3P2}
\end{figure}
If the datasets are $\{X_1, X_2, X_4\}$ and $\{X_1, X_3, X_4\}$, there is no information about conditional independence available to use.  Assuming Gaussian distributions, the unified graph we get from those two marginal datasets is figure \ref{fig:Case3P2U}:
\begin{figure}[H]
    \centering
    \includegraphics[width=100mm]{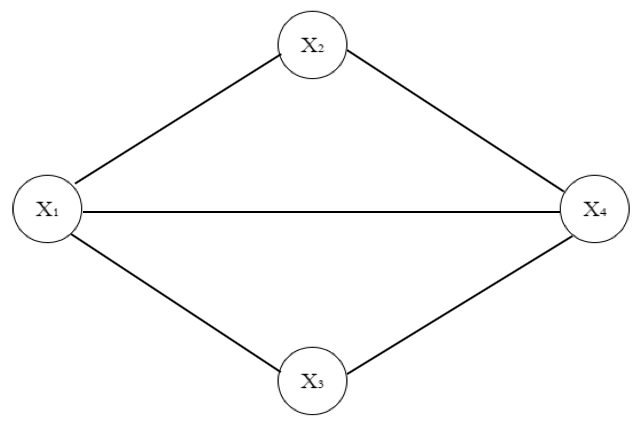}
    \caption{Unified Graph with Gaussianity}
    \label{fig:Case3P2U}
\end{figure}
Namely, each marginal dataset tells us that every pair of variables are dependent, so what we get are two triangles.  To estimate a unified causal graph, we can only put the two triangles together.

However, if we assume non-Gaussian distributions, we can run causal discovery algorithms working on non-Gaussian distributions, such as LiNGAM, on each dataset.  Such algorithms will estimate  the direction of inference between each pair of variables.  For figure 2 with marginal datasets $\{X_1, X_2, X_4\}$ and $\{X_1, X_3, X_4\}$, we get the two graphs below:
\begin{figure}[H]
    \centering
    \includegraphics[width=100mm]{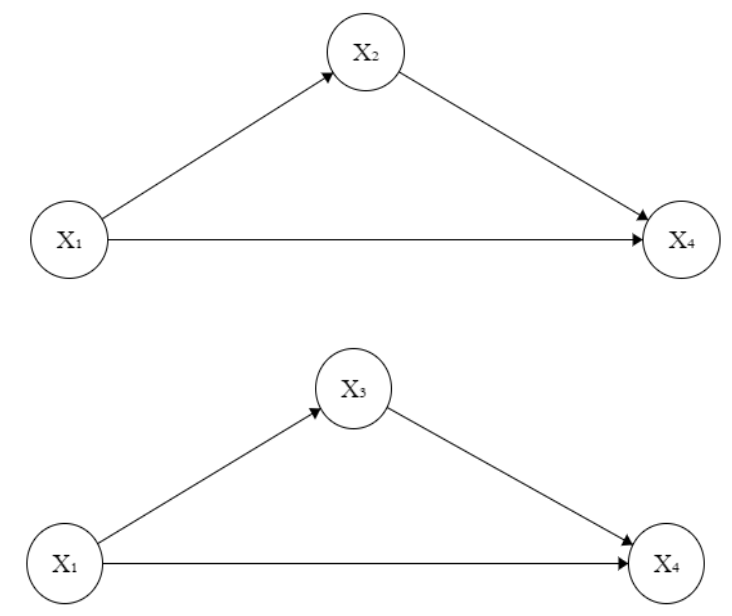}
    \caption{Two Possible Graphs with non-Gaussianity}
    \label{fig:Case2Tr}
\end{figure}
Note that now since we know the direction of each edge, an undirected trek (including $X_1, X_2, X_3$ and $ X_4$) between $X_1$ and $X_4$ can be identified.  If there exist other treks connecting $X_1$ and $X_4$, by the TREK rule, we should have:
\begin{center}
    $\rho_{14}>a_{12}a_{24}+a_{13}a_{34}  or  \rho_{14}<a_{12}a_{24}+a_{13}a_{34}$
\end{center}
Since every variable is standardized, from the two triangles above we know that the coefficient on each edge equals the correlation between two variables connected by that edge (if there are no other treks between those variables).  In order to check whether the inequality holds, we just need to plug the corresponding correlations into the formula.  If instead of an inequality, what we get is an equality:
\begin{center}
    $\rho_{14}=a_{12}a_{24}+a_{13}a_{34}$
\end{center}
We can conclude that there is no other trek connecting $X_1$ and $X_4$ and get the true causal graph by removing the edge between them.

\section{ General Principles}
So far examples above shows that using the TREK rule to estimate causal connection follows these principles:
\begin{enumerate}
    \item Possible treks can be identified by conditional independence;
    \item 	Comparing the absolute value of correlation between variables in the same trek rules out candidate causal graphs;
    \item Calculating correlations by the TREK rule on possible treks rules out redundant connections between variables.
\end{enumerate}
The three principles above generally depict how the TREK rule works:  in order to apply the rule, the first step is to determine which two or more variables are potentially connected by treks and what variables are contained in the trek, which is principle 1; after identifying a potential trek and its component, we can compare the absolute value of correlations between variables in the same trek and rule out all the causal graphs that violates the TREK rule based on the result of the comparison, which is principle 2;  furthermore, if available correlation allows, we can calculate the theoretical correlation between variables being connected by treks and use it to estimate the existence of latent variable or omitted direct connection, which is the principle 3 and is used in case three.

\section{ TREK Rules Can Inform the Choice of Further Experiments}
All those cases provided in the last three sections show that instead of just enumerating all possible unified causal graphs consistent with the conditional independence revealed by marginal datasets, for linear systems we can make more specific estimations about connections between variables by applying the TREK rule, such as removing redundant edges or determining relative positions of variables in a path.  Furthermore, the motivation of applying the TREK rule can guide researchers to make future measurements more efficient.  The idea is illustrated in the case below:
\begin{figure}[H]
    \centering
    \includegraphics[width=100mm]{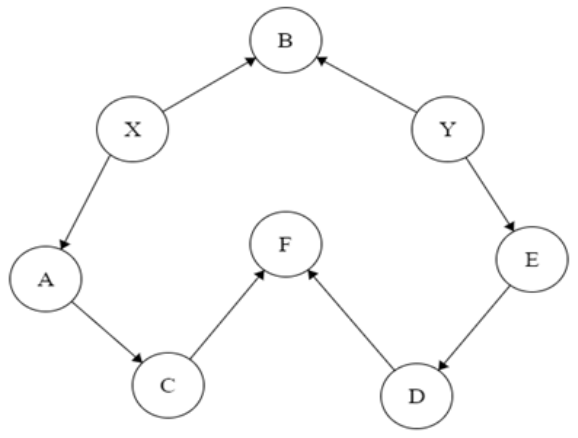}
    \caption{True Graph}
    \label{fig:8M}
\end{figure}
Consider a situation where the true graph is figure \ref{fig:8M}.  Suppose that for all these variables that researchers are interested in, only a few of them can be measured together each time.  Now consider that currently available marginal datasets are:
\begin{center}
    $\{X, Y, A\}$; $\{X, Y, B\}$; $\{X, Y, C\}$; $\{X, Y, D\}$; $\{X, Y, E\}$; $\{X, Y, F\}$
\end{center}

The information about conditional independence and dependence we can get from these datasets is limited: from $\{X, Y, B\}$ and $\{X, Y, F\}$, we know that B and F are colliders between $X$ and $Y$; from other datasets we only get that marginal dependence relations and independence relations between X or Y and other variables (for instance, from$\{X, Y, A\}$ we get $X \indep Y$ and $A \indep Y$).  Even if we assume non-Gaussian distributions and know the direction of edges between variables dependent on each other, such as $X\rightarrow A$ and $Y\rightarrow D$, the TREK rule cannot be usefully applied; there are too many candidates of unified causal graphs that satisfy these dependence and independence relations.
However, we can observe that 
\begin{center}
    $|\rho_{XA}| > |\rho_{XC}| > |\rho_{XF}|$
\end{center}

Potentially there is a TREK connecting X and F that contains $A$ and $C$.  The existence of such a trek cannot be verified directly because we do not know $|\rho_{AC}|$ and $|\rho_{FC}|$, but if future measurements are possible, we can make one more measurement: $\{B, F, C\}$.  In this way, we can get three more correlations: $|\rho_{BC}|$, $|\rho_{FC}|$ and $|\rho_{FB}|$.
Now we can test whether $A$ and $C$ are in the trek connecting $X$ and $F$.  If such a trek exists and there are no other treks connecting $X$ and $F$, then we should have:
\begin{center}
    $|\rho_{XC}|=|\rho_{AC}||\rho_{AX}|$
\end{center}
and
\begin{center}
    $|\rho_{XF}|=|\rho_{AX}||\rho_{AC}||\rho_{CF}|$\footnote{Here we are using $\rho$ (correlation) and a (coefficient) interchangeably. It is because here we are testing if $A$ and $C$ are in the unique trek connecting $X$ and $F$.  If $A$ and $C$ are in the unique trek connecting $X$ and $F$, then we should have $\rho_{XA}= a_{XA}$, $\rho_{AC}=a_{AC}$, $\rho_{CF}=a_{CF}$.}
\end{center}
Therefore, if we find
\begin{center}
    $\dfrac{|\rho_{XC}|}{|\rho_{XA}|} = \dfrac{|\rho_{XF}|}{|\rho_{XA}||\rho_{CF}|}$
\end{center}
then it is likely that $A$ and $C$ are in the trek connecting $X$ and $F$.

Moreover, we can also find:
\begin{center}
    $|\rho_{BF}|= |\rho_{XB}||\rho_{XF}|+|\rho_{YB}||\rho_{YF}|$   
\end{center}
                             
From equation above we can conclude that $B$ and $F$ are connected by two treks: one of which contains $X$, $A$, and $C$, and can be fully determined; the other contains $Y$.  Since can we observe that: 
\begin{center}
    $|\rho_{YE}| > |\rho_{YD}| > |\rho_{YF}|$
\end{center}

It is possible that $E$ and $D$ are in the trek connecting $B$ and $F$ which has $Y$.  We can estimate whether:
\begin{center}
    $\dfrac{|\rho_{BF}| - |\rho_{XB}||\rho_{XF}|}{ |\rho_{BY}||\rho_{YE}|} =\dfrac{|\rho_{FY}|}{|\rho_{FE}|}\footnote{Again, we are using correlation and coefficient interchangeably} $
\end{center}

By the TREK rule, if this equation holds, then it is likely that $E$ is in the other trek connecting $B$ and $F$.  Based on all these conclusions, we nearly recover the true graph.

Notice that here the trek connecting $B$ and $F$ is the longest trek that could exist given the initial pack of marginal datasets.  Measuring B and F together with an additional variable could enable us to use the TREK rule involving more variables and get much more information than measuring other variables together.

From this case, we can see that although most of time the TREK rule cannot identify a unique unified causal graph (which is highly dependent on what marginal datasets are available), it can be helpful as a criterion to plan future measurements. 
\section{ Discussion}
The limitation of the TREK rule to linear systems is less stringent than it may appear. Non-linear systems can be transformed into linear systems in several ways that preserve the graphical causal structure.  One long-standing method is domain specific transformations of individual variables. Econometric models, for example commonly express prices as logarithms, presumably because economists decided long ago that the log of prices has a Normal (Gaussian) distribution. But there are more general, domain independent transformations. For any of a large family of probability distributions (roughly, those whose cumulative distribution function has a smooth, monotonic map to the cumulative distribution of the Gaussian) a nonparanormal transformation yields a joint Gaussian distribution \cite{Liu:2009:NSE:1577069.1755863}.  The relations among the variables can be expressed as linear regressions in these transformed variables, with additive disturbances. The regression coefficients obey the TREK rules when the transformed variables are standardized.

However, the TREK rule is not practical for dense graphs in which a pair of variables is connected by several treks.  When the graph is dense the choices of marginal data sets will interact with trek rules in complex ways that may prevent obtaining useful information from trek constraints.

\def\bibfont{\small}
\bibliographystyle{plain}
\bibliography{trekrule}

\end{document}